\RequirePackage{fix-cm}
\documentclass[smallextended]{svjour3}       
\smartqed  
\usepackage{graphicx}
\usepackage{subfigure}
\usepackage{Tabbing}
\usepackage{multirow}
\usepackage[round]{natbib}
\usepackage{amsmath}
\usepackage[boxed,vlined,linesnumbered]{algorithm2e}
\usepackage{url}
\usepackage{amssymb}

\makeatletter
\newenvironment{Ualgorithm}[1][htpb]
  {\def\@algocf@post@ruled{\kern\interspacealgoruled\hrule  height\algoheightrule\kern3pt\relax}%
    \def\@algocf@capt@ruled{under}
    \begin{algorithm}[#1]}
  {\end{algorithm}}
\makeatother

\begin{document}

\title{Measuring Divergence in Dependency Trees
       to Improve Projection Algorithms
}
\subtitle{}


\author{Ryan Georgi         \and
        Fei Xia \and
        \hbox to 85pt{\hss William D. Lewis\hss}  
}


\institute{R. Georgi \\ F. Xia \at
              University of Washington Department of Linguistics \\
              Box 354340 Seattle, WA 98195-4340 \\
              Tel.: +1 (206) 543-2046\\
              Fax: +1 (206) 685-7978\\
              \email{rgeorgi@uw.edu}           \\
              \email{fxia@uw.edu}
           \and
           W. D. Lewis \at
           	Microsoft Research, Bldg 99\\
			14820 NE 36th St, Redmond, WA 98052-6399 \\
			\email{wilewis@microsoft.com}
}

\date{Received: date / Accepted: date}

\maketitle

\begin{abstract}
Obtaining syntactic parses is a crucial part of many NLP pipelines. However, most of the world's languages do not have large amounts of syntactically annotated corpora available for building parsers. Syntactic projection techniques attempt to address this issue by using parallel corpora consisting of resource-poor and resource-rich language pairs, taking advantage of a parser for the resource-rich language and word alignment between the languages to project the parses onto the data for the resource-poor language. These projection methods can suffer, however, when the two languages are divergent.
In this paper, we investigate the possibility of using small, parallel, annotated corpora to automatically detect divergent structural patterns between two languages. These patterns can then be used to improve structural projection algorithms, allowing for better performing NLP tools for resource-poor languages, in particular those that may not have large amounts of annotated data necessary for traditional, fully-supervised methods. While this detection process is not exhaustive, we demonstrate that common patterns of divergence can be identified automatically without prior knowledge of a given language pair, and the patterns can be used to improve performance of projection algorithms.
\end{abstract}

\section{Introduction}
\label{intro}

When it comes to resources for natural language processing, a small handful of languages account for the vast majority of available resources. Out of the resources listed by the LRE Map \citep{Calzolari:2012vn}, English accounts for 30\% of all recorded resources, and the ten most resourced languages for 62\% of all resources. A broad variety of tools are available for these resource-rich languages, as the time and effort spent to annotate resources for these languages allows for state-of-the-art systems to be built utilizing supervised and semi-supervised methods.

	The availability of such resources is the result of a large investment over many years on a per-language-basis. Because creating high-quality annotation is expensive and labor intensive, the vast majority of the world's languages lack such resources and high-performance NLP tools. To address this issue, recent studies \citep{Lewis:2008va, Benajiba:2010fk, Georgi:2012ys} have proposed to take advantage of bitext and resources for resource-rich languages; that is, use tools for resource-rich languages to process one side of bitext (the resource-rich language) and project the information to the other side of bitext (the resource-poor language) via word alignments.
	
	While projecting annotation from one language to another is a promising method for adding annotation to languages using automated methods, it relies on the assumption that simple word alignments between languages are sufficient to represent analogous meanings and structures between the languages. For reasons we will discuss in the following sections, this assumption is useful, but often erroneous.

	Finding out whether and when this assumption fails for a given language pair is not easy without knowledge about the two languages. It would be useful if, given a small set of seed data, a language pair could be analyzed to find where and in what ways the languages diverge, and use these detected patterns as corrective guidelines for performing projection upon other sentences of the language pair.

In this paper, we propose a method for analyzing a language pair and determining the degree and types of divergence between two languages.
This systematic identification of divergence types could then lead to better informed syntactic projections, and subsequently can improve the tools
built upon such data.

\section{Background}
\label{sec:previous}

While there is a growing body of work on projection methods as a means to bootstrap resources from one language to another, there are not many studies on how to treat the issue of linguistic divergence between these languages. In this section, we provide a brief review of work on divergence and projection algorithms. We will also introduce interlinear glossed text (IGT), a common format for linguists to represent language examples.

\subsection{Projection Methods} 

\begin{figure}
\center
\includegraphics[width=0.55\textwidth]{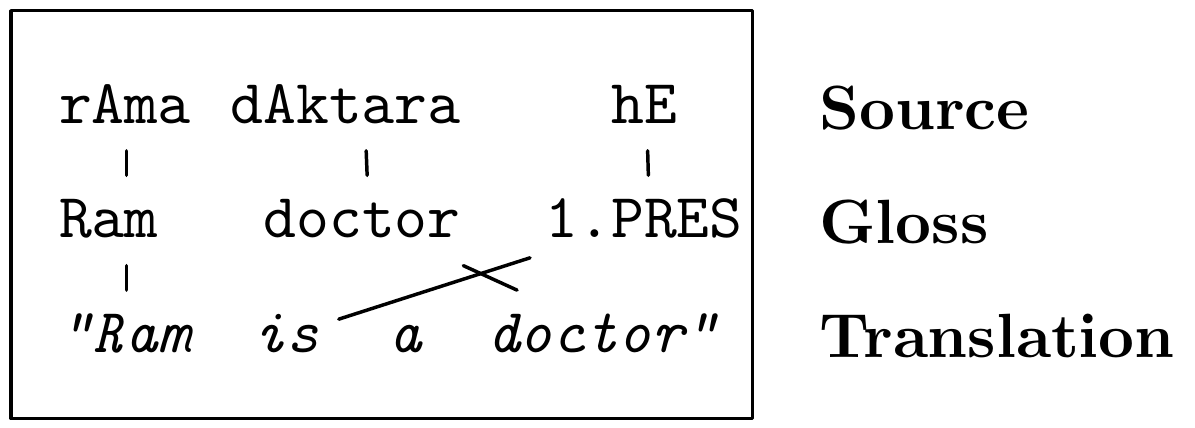}
\caption{An instance of Interlinear Glossed Text (IGT) for Hindi, with a gloss line and translation in English. The word alignment between the gloss and translation lines is not part of the IGT instance, but it can be produced by a statistical word aligner trained on bitext or an aligner that uses heuristic rules.}
\label{fig:igt}
\end{figure}

\newcommand{\projectwidth}{0.7\textwidth}
\begin{figure}
\centering
\subfigure[Using the interlinear instance from Figure \ref{fig:igt}, the English text is parsed.]{\includegraphics[width=\projectwidth]{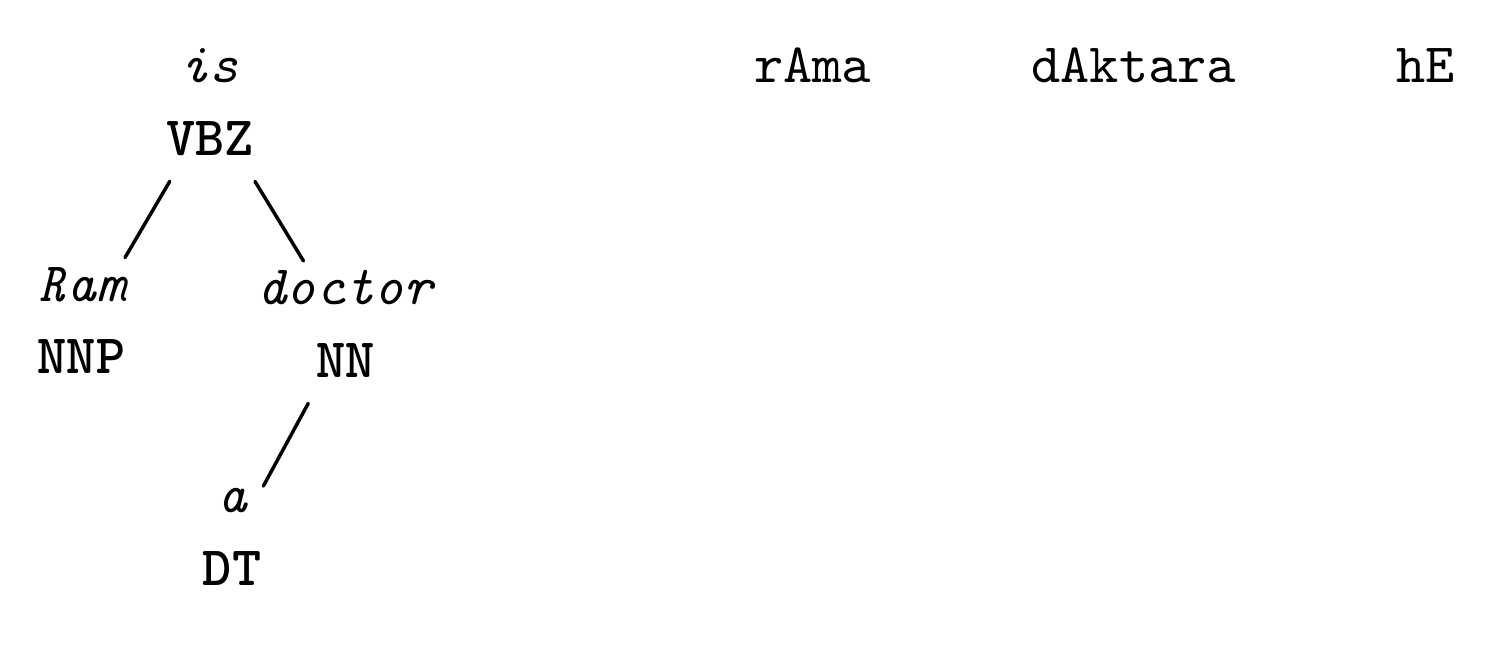}}
\subfigure[Using the word alignments from Figure \ref{fig:igt}, the tree structure and POS tags for the English tree are ``projected'' onto the Hindi sentence.]{\includegraphics[width=\projectwidth]{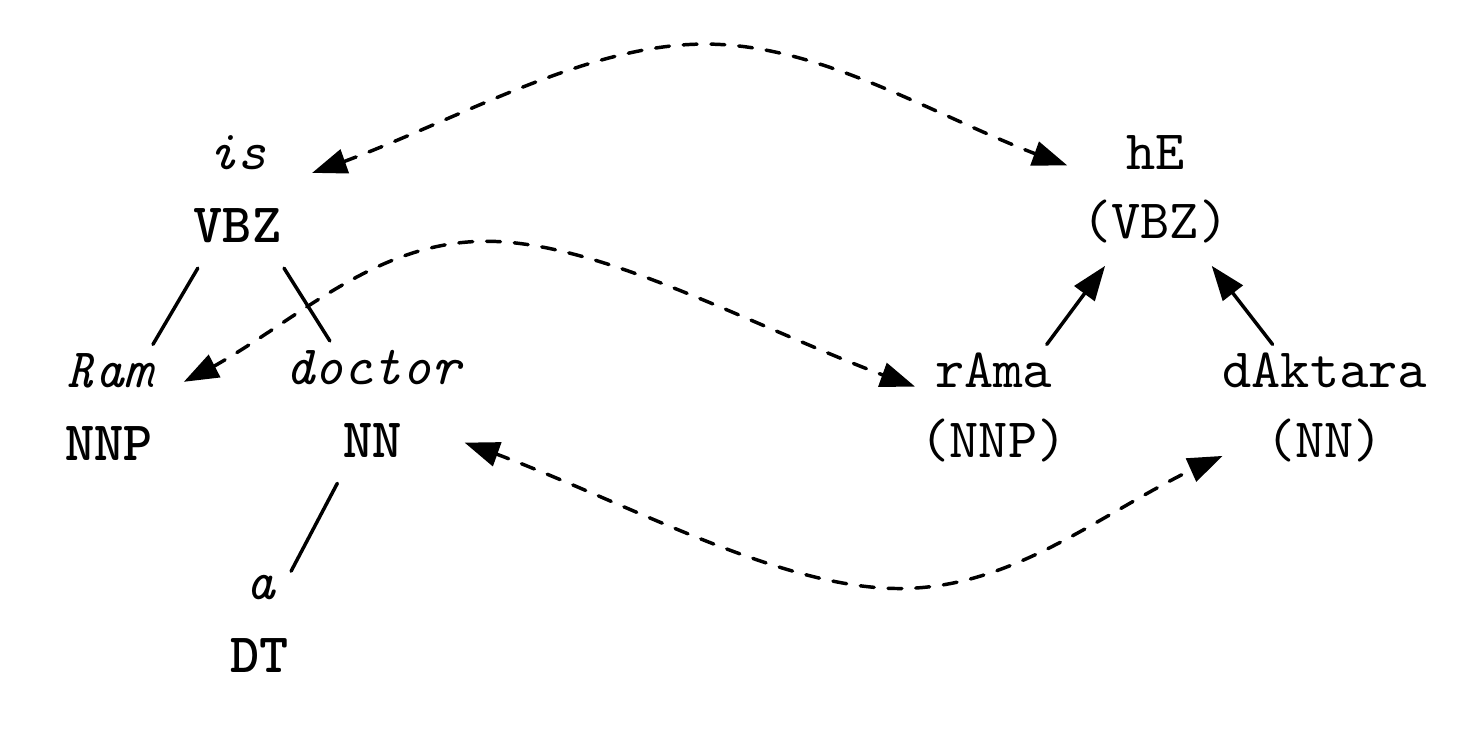}}
\caption[Simplified syntactic projection example]{A simple example of syntactic projection as performed on the IGT instance in Figure \ref{fig:igt}.}
\label{fig:projection}
\end{figure}

Projection algorithms have been the target of a fair amount of research in the last decade as attempts have been made to utilize statistical alignment methods to match words between languages with parallel data and ``project'' annotation between the two. Figure \ref{fig:igt} shows an example bitext in the form of an interlinear glossed text (IGT), while Figure \ref{fig:projection} shows how this bitext may be used to project a dependency tree from the English to the Hindi.

Some of the initial research on the subject of projecting word-level annotation from one language to another was published in \cite{Yarowsky:2001dl}. Here, the authors used IBM Model 3 \citep{Brown:1990zr} to align large parallel corpora between English--Chinese and English--French. A POS tagger was trained for French using projection from English, and NP bracketers were trained similarly for both French and Chinese. The authors identified  noisy statistical alignment and 1-to-many alignments as two main issues in performing projection. The first of these issues is indeed a difficult problem for resource-poor languages, as high-quality statistical word alignment often requires much more bitext than the data available for the language. While it is not a full solution to the problem, many of the language pairs we use in this work are drawn from collection of interlinear glossed text (IGT),  as shown in Figure \ref{fig:igt}, which provides unique shortcuts for obtaining word alignment with a small amount of data. IGT will be discussed further in \S\ref{sec:igt}.

\begin{figure}
\center
\includegraphics[width=0.35\textwidth]{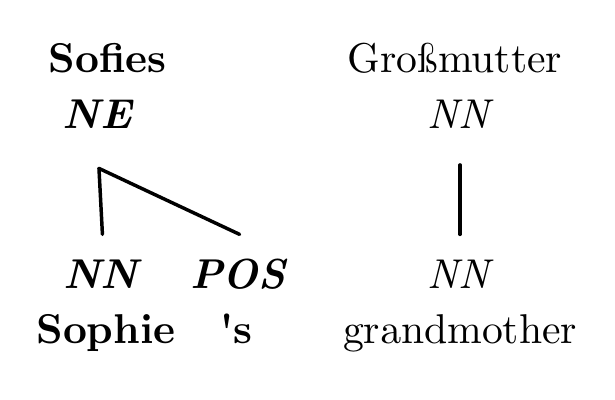}
\caption{Simple but frequent example of 1-to-many German--English alignment found in the {\it Sophie's World} portion of the SMULTRON corpus \citep{Smultron2010}.}
\label{fig:sophie}
\end{figure}

The second issue of 1-to-many alignments is one that may be the result of linguistic divergence between a language pair where the source is morphologically richer than the target. In cases such as this, finding common patterns of conflation can be useful for generalizing a projection to new data. For instance, Fig. \ref{fig:sophie} shows a very simple but common case of conflation in the SMULTRON corpus \citep{Smultron2010} where a single German word aligns to multiple English words. Using direct projection alone, the same POS tag would be projected to both English tokens. In this case, using a universal tagset such as those presented by \cite{Petrov:2012ly} could help alleviate the problem, but for more complex cases, learning the pattern would be more critical.

 \cite{Hwa:2004ua} investigate the issues in using projection between languages to develop and train syntactic parsers, and outline the Direct Correspondence Assumption (DCA), the assumption made in projection algorithms that the target language tree should be homomorphic to the source language tree. While useful, this assumption often does not hold, as the authors point out. In order to fix some of the errors made by faulty projections,  \citeauthor{Hwa:2004ua} use an approach that applies post-projection correction rules. For projection from English to Spanish, the accuracy of the projected structures increased from 36.8\% to 70.3\%. The accuracy of the English to Chinese projection increased from 38.1\% and 67.3\%.
 
While these language-specific rewrite rules are promising, they still require language-specific knowledge. What we seek to accomplish in this paper is a general framework for automatically detecting this divergence, both in specific language pairs, and its frequency throughout a large number of languages. With the use of this automated divergence detection, it may be possible to learn these rewrite rules from a small annotated corpus and use them to improve projection algorithms.

\subsection{Interlinear Glossed Text}
\label{sec:igt}

As mentioned in the preceding section, much of the data for our experiments was drawn from the unique IGT data type. IGT instances are a common way for linguists to give illustrative examples for a language being studied.  Figure \ref{fig:igt} shows an instance of IGT for Hindi and English. Of special interest in IGT instances is the middle gloss line which gives a word-by-word gloss of the source language. This annotated gloss typically contains words from the English translation, albeit reordered and with extra morphological and semantic information. The matching of gloss and translation can be utilized to obtain high-quality automatic word alignment between source and target, without the need for far larger amounts of data typically required by statistical alignment algorithms.

In \cite{Xia:2007wv}, enriched IGT data for 7 language pairs was created using this augmented alignment and structural projection, then hand-corrected to create gold standards with minimal expert intervention.  They showed the potential for using IGT as a resource for languages for which finding resources would otherwise be extremely difficult or impossible to obtain.  We will use this data for the current work. A breakdown of the language pairs can be seen in \S\ref{sec:data}.

\cite{Lewis:2008va} used projected phrase structures 
to determine the basic word order for 97 languages using
a database of IGT instances. By using the alignment method described above and projecting phrase structures from English to the language line, the word order in the foreign language could be inferred.
For languages with just 10--39 IGT instances,
the accuracy of predicting basic word order was 79\%; with more than 40 instances, the accuracy jumped to
99\%.

\subsection{Linguistic Divergence}

These studies illustrate the promise of projection for bootstrapping new tools for resource-poor languages, but one limitation is their reliance on the assumption
that syntactic structures of the two sentences
in a given sentence pair are similar. While \citeauthor{Hwa:2002wp}'s  
{\it Direct Correspondence Assumption} (DCA) describes the assumption made for projection,  \cite{Dorr:1994:MTD:203987.203993} makes a deeper analysis of divergence in languages. \citeauthor{Dorr:1994:MTD:203987.203993} outlines {\it lexical conceptual structures} (LCS) that provide a general framework to describe these exceptions to the DCA. This framework is capable of representing divergence stemming from syntactic, lexical, or semantic differences, but for the purposes of this paper we will focus primarily on those that are lexical and syntactic.

 \citeauthor{Dorr:1994:MTD:203987.203993} identifies a number of ways in which languages may diverge, specifically syntactic and semantic differences in mappings between languages. Our goal in this work is to create a methodology by which
some common types of divergences can be detected automatically 
from bitexts in order to improve the performance of existing structural projection methods.

\section{Methodology}

In our approach to automatically detecting divergent structures between language pairs, we first propose a metric to measure the
degree of {\it matched} edges between source and target trees (\S\ref{sec:comparing}).
Second, we define three operations on trees in order to capture
three common types of divergence (\S\ref{sec:operations}).
Third, we apply the operations on a tree pair and show how
the operations could affect the degree of tree {\it match}
(\S\ref{sec:apply-operations}).
 After explaining the relationship of our operations to \citeauthor{Dorr:1994:MTD:203987.203993}'s divergence types (\S\ref{sec:dorrtypes}), we discuss how knowing these divergence types can be useful in improving structural projection algorithms (\S\ref{sec:improve-projection}).

\subsection{Comparing Dependency Trees}
\label{sec:comparing}

\newcommand{\subfigwidth}{0.3\columnwidth}
\begin{figure}
\center
\subfigure[A {\it match} alignment]{\includegraphics[width=\subfigwidth]{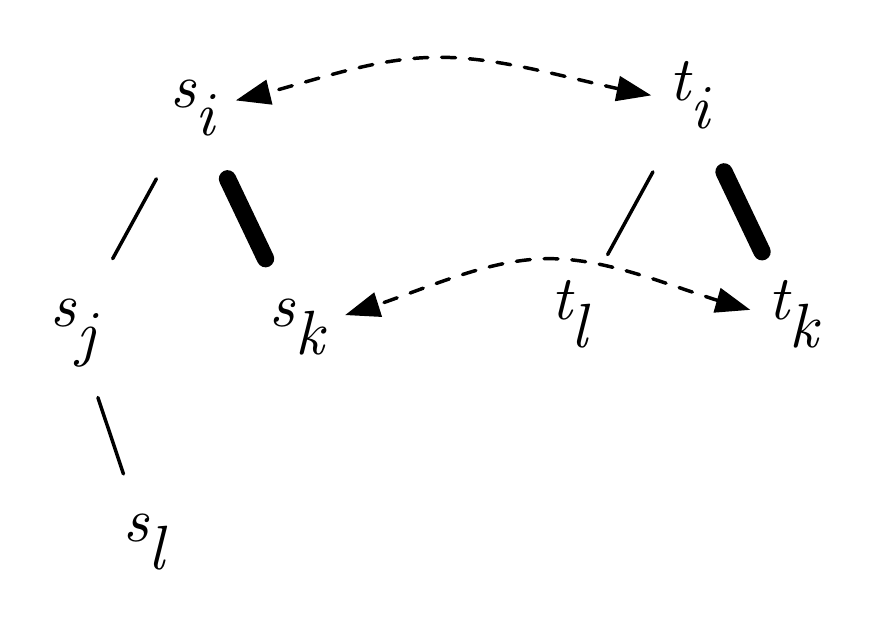}}
\subfigure[A {\it merge} alignment]{\includegraphics[width=\subfigwidth]{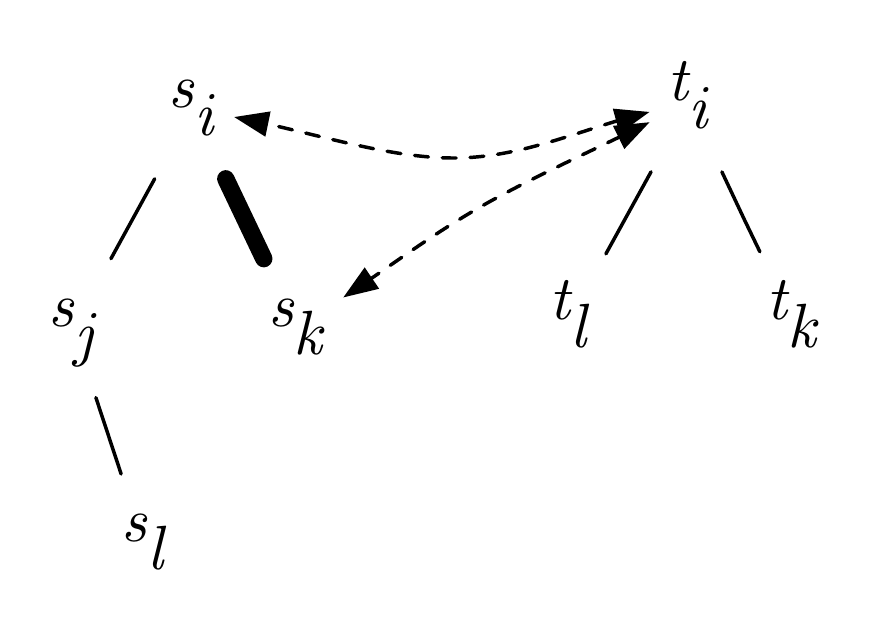}\label{fig:merge-align}}
\subfigure[A {\it swap} alignment]{\includegraphics[width=\subfigwidth]{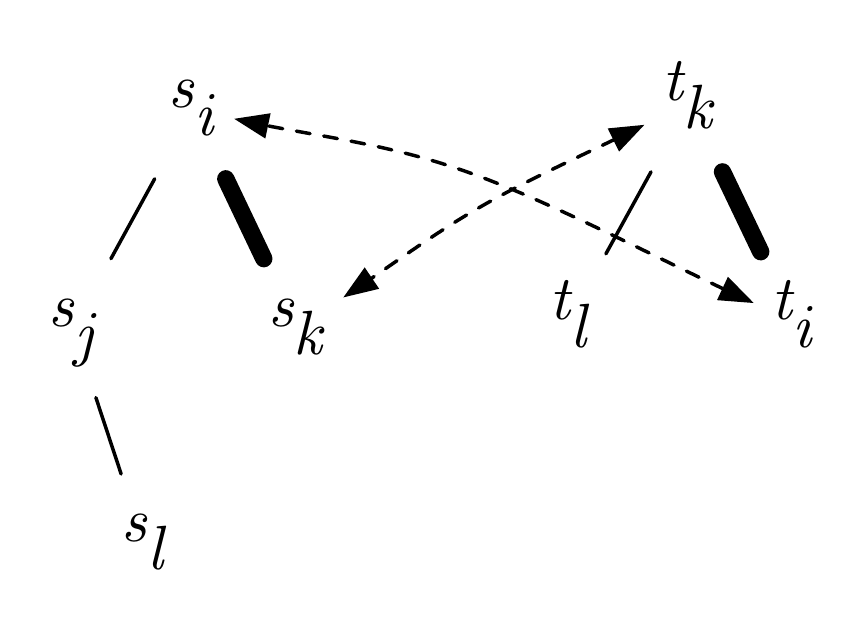}\label{fig:swap-align}}
\caption{Definition of a {\it match}, {\it merge}, and {\it swap} edges in a tree pair.}
\label{fig:match-align}
\end{figure}

One of the key aspects of our method was devising a metric to compare dependency trees cross-linguistically, as most existing tree similarity measures are intended to compare tree representations with the same number of tokens. Comparing between languages, on the other hand, means that the number of tokens can vary. We instead look for a method to determine similarity by means of matched edges in the tree, as shown in Figure \ref{fig:match-align}.

Given an IGT, let $F$ be the parse tree for the language line (aka source tree), $E$ be the parse tree for the translation line (aka target tree), and $A$ be the word alignment between the language line and the translation line. $F$ is made up of the words $W$, in the language line, and a set of edges between the words, as follows:

\begin{align}
& F = (W_F, T_F)\\
& W_F = (f_1 \ldots f_n)\\
& T_F = \big\{(f_i, f_k) \ldots (f_n, f_m)\big\}
\end{align}

\noindent E is defined similarly, except words in the translation line are denoted as $e_i$, not $f_i$. The alignment $A$ is a set of word pairs:

\begin{equation}A = \big\{(f_i, e_k) \ldots (f_j, e_l)\big\}\end{equation}

We call an $(F, E, A)$ tuple an aligned tree pair. A Corpus, $C$, in our experiments, is a set of $(F,E,A)$ tuples.

An edge $(f_i, f_k)$ in the 
foreign-language tree  
is said to match an edge $(e_i, e_k)$ in the english-language tree if $f_i$ is aligned to $e_i$ and $f_k$ is aligned to $e_k$. Because the alignment between a sentence pair can be many-to-many, we define the following functions, which map a word from one sentence to the set of words in the other sentence.

\begin{align}
& R_{F\rightarrow E}(f_i, A) = \Big\{e | (f_i, e) \in A \Big\} \\
& R_{E\rightarrow F}(e_i, A) = \Big\{f | (f, e_i) \in A \Big\}
 \end{align}

\noindent We then define the boolean function $match$, as follows:

\begin{align}
 match(f_i, f_j, T_E, A) = \left \{  \begin{array}{rl}
 1 & {\bf if }\  \exists e_a, e_b \Bigg(\Big(e_a \in R_{F\rightarrow E}(f_i)\Big) \wedge \\ 
 & \hspace{15mm} \Big( e_b \in R_{F\rightarrow E}(f_j) \Big) \wedge \Big( (e_a, e_b) \in T_E\Big)\Bigg)  \\ 
 0 & otherwise \\
 \end{array}
 \right.
\end{align}

That is, an edge $(f_i, f_j)$ in $F$ matches some edge in $E$ according to $A$ if there exists two target words, $e_a$ and $e_b$ in $E$ such that $e_a$ aligns to $f_i$, $e_b$ aligns to $f_j$, and $(e_a, e_b)$ is an edge in $E$.

Given an aligned tree pair $(F,E,A)$, we define $SentMatch(F,E,A)$ as the percentage of edges in $F$ that match some edge in $E$. Given a corpus $C$, we define $CorpusMatch_{Src\rightarrow Tgt}(C)$ as the percentage of edges in the source trees that match some edges in the corresponding target trees. Similarly, $CorpusMatch_{Tgt\rightarrow Src}(C)$ is the percentage of edges in the target trees that match some edges in the corresponding source trees.

\begin{align}
CorpusMatch_{Src\rightarrow Tgt}(C) = \frac{\displaystyle \sum_{(F,E,A)\ \in\ C}\left({\sum_{(f_{i},f_{j})\ \in\ T_{F}} match(f_i, f_j, T_E, A)}\right)}
{\displaystyle \sum_{(F,E,A) \in C} \big|T_{F}\big|}
\end{align}

\subsection{Defining Tree Operations}
\label{sec:operations}

 When an edge $(f_i, f_k)$ in a tree does not match any edge in the other tree, it may be caused by one of the following cases:
	\renewcommand{\labelenumi}{C\arabic{enumi}.}
	\begin{enumerate}
	\item $f_i$ or $f_k$ are spontaneous (they do not align with any words in the other tree). \label{case1}
	\item $f_i$ and $f_k$ are both aligned with the same node $e_i$ in the other tree  (Fig \ref{fig:merge-align}). \label{case2}
	\item $f_i$ and  $f_k$ are both aligned with nodes in the other tree, $e_k$ and $e_i$, but in a reversed parent-child relationship (Fig \ref{fig:swap-align}). \label{case3}
	\item There is some other structural differences not caused by C\ref{case1}--C\ref{case3}. \label{case4}
	\end{enumerate}
	
	The first three cases are common. To capture them, we define three operations on a tree --- \textit{remove}, \textit{merge}, and \textit{swap}.

\subsubsection*{O1 -- Remove}

The \textit{remove} operation is used to remove spontaneous words. As shown in Figure \ref{fig:o1}, removal of the node \textit{l} is accomplished by removing the link between node \textit{l} and its parent, \textit{j}, and adding links between the parent and the removed node's children.

This result of this operation looks can be seen in Figure \ref{fig:o1}, using the relation $Children$, which maps a word to the set of all its children in the tree.

\RestyleAlgo{ruled}
\begin{Ualgorithm}[h]
\scriptsize
\TitleOfAlgo{$Remove(w, T)$}
\SetKwInOut{Output}{Output}
\SetKwInOut{Input}{Input}
\Input{$T = \big\{(w_i, w_j) \ldots (w_m, w_n)\big\}$ \tcp*{Input tree }}
\Input{$w$ \tcp*{Word to remove.}}
\Output{$T^{\prime}$ \tcp*{Modified tree}}
\Begin{
	$T^{\prime} = T - \big\{(w, w_i) | w_i = parent(w, T)\big\}$ 
				\tcp*[f]{Remove edge between $w$ and parent $w_i$}\\
							\hspace{10.5mm} $ - \big\{(w_j, w) | w = parent(w_j, T)\big\}$
							\tcp*[f]{Remove edges for children of $w$}\\
							\hspace{10.5mm} $ + \big\{(w_j, w_i) | w_i=parent(w, T), w = parent(w_j, T)\big\}$
							\;
				\BlankLine
				\tcc{Finish by ``promoting'' former children of $w$ to now attach to $w$'s parent, $w_i$.}
				\BlankLine
}
\Return $T^{\prime}$
\caption[{\it Remove} operation]{Remove a token $w$ from the tree $T$.}
\label{alg:remove}
\end{Ualgorithm}

\subsubsection*{O2 -- Merge}

The \textit{merge} operation is used when a node and some or all of its children in one tree align to the same node(s) in the other tree, as can be seen  in Figure \ref{fig:merge-align}. The parent \textit{j} and child \textit{l} are collapsed into a merged node, as indicated by \textit{l+j} in Figure \ref{fig:o2}, and the children of \textit{l} are promoted to become children of the new node \textit{l+j}. The result can be seen in Figure \ref{fig:o2}.

\begin{Ualgorithm}[h!]
\scriptsize
\TitleOfAlgo{$Merge(w_c, w_p, T)$}
\SetKwInOut{Output}{Output}
\SetKwInOut{Input}{Input}
\Input{$T = \big\{(w_i, w_j) \ldots (w_m, w_n)\big\}$ \tcp*{Input tree }}
\Input{$w_c$ \tcp*{Child word to merge.}}
\Input{$w_p$ \tcp*{Parent word to merge.}}
\Output{$T^{\prime}$ \tcp*{Modified tree}}
\Begin{
	$T^{\prime} = T -  \big\{ (w_c, w_p) \big\}$\\
							$\hspace{14mm} - \big\{(w_i, w_c) | w_c = parent(w_i, T)\big\}$\\
							\hspace{13mm} $ + \big\{(w_i, w_p)|w_c = parent(w_i, T)\big\}$\;
}
\Return $T^{\prime}$
\caption[{\it Merge} operation]{Merge a child $w_c$ and parent $w_p$ in the tree $T$, and ``promote'' the children of $w_c$ to be children of $w_p$.}
\label{alg:merge}
\end{Ualgorithm}

\subsubsection*{O3 -- Swap}

The \textit{swap} operation is used when two nodes in one tree are aligned to two nodes in the other tree, but in a reciprocal relationship, as shown in Figure \ref{fig:swap-align}. This operation can be used to handle
certain divergence types such as
\textit{demotional} and \textit{promotional} divergence, which will be discussed in more detail in \S\ref{sec:dorrtypes}.

Figure \ref{fig:o3} illustrates how the swap operation takes place by swapping nodes \textit{l} and \textit{j}. Node \textit{j}, the former parent, is \textit{demoted}, keeping its attachment to its children. Node \textit{l}, the former child, is \textit{promoted}, and its children become siblings of node \textit{j}, the result of which can be seen in Figure \ref{fig:o3}.

\begin{Ualgorithm}[h!]
\scriptsize
\TitleOfAlgo{$Swap(w_c, w_p, T)$}
\SetKwInOut{Output}{Output}
\SetKwInOut{Input}{Input}
\Input{$T_L = \big\{(w_i, w_j) \ldots (w_m, w_n)\big\}$ \tcp*{Input tree }}
\Input{$w_c$ \tcp*{Child word to swap.}}
\Input{$w_p$ \tcp*{Parent word to swap.}}
\Output{$T^{\prime}$ \tcp*{Modified tree}}
\Begin{
	$T^{\prime} = T -  \big\{ (w_c, w_p) \big\} + \big\{(w_p, w_c)\big\}$
							\tcp*[f]{Swap the order in the $(w_c, w_p)$ edge} \\
							$\hspace{14mm} - \big\{(w_p, w_i) | w_i = parent(w_p, T)\big\}$
							\tcp*[f]{Remove edges for parent $w_p$}\\
							\hspace{13mm} $ + \big\{(w_c, w_i) | w_i = parent(w_p, T)\big\}$
							\tcp*{Add edges from $w_p$'s parent to $w_c$}
}
\Return $T_L^{\prime}$
\caption[{\it Swap} operation]{Swap a child $w_c$ and parent $w_p$ in the tree $T$.}
\label{alg:swap}
\end{Ualgorithm}

\newcommand{\examplewidth}{0.45\textwidth}

\begin{figure}
\center
\subfigure[Before and after the node $l$ has been removed (O1).]{
\raisebox{16mm}{$Remove(l, T): \left \{ \begin{array}{l} Children(j) = \{o, p, m, n\} \end{array} \right.$}
\includegraphics[width=\examplewidth]{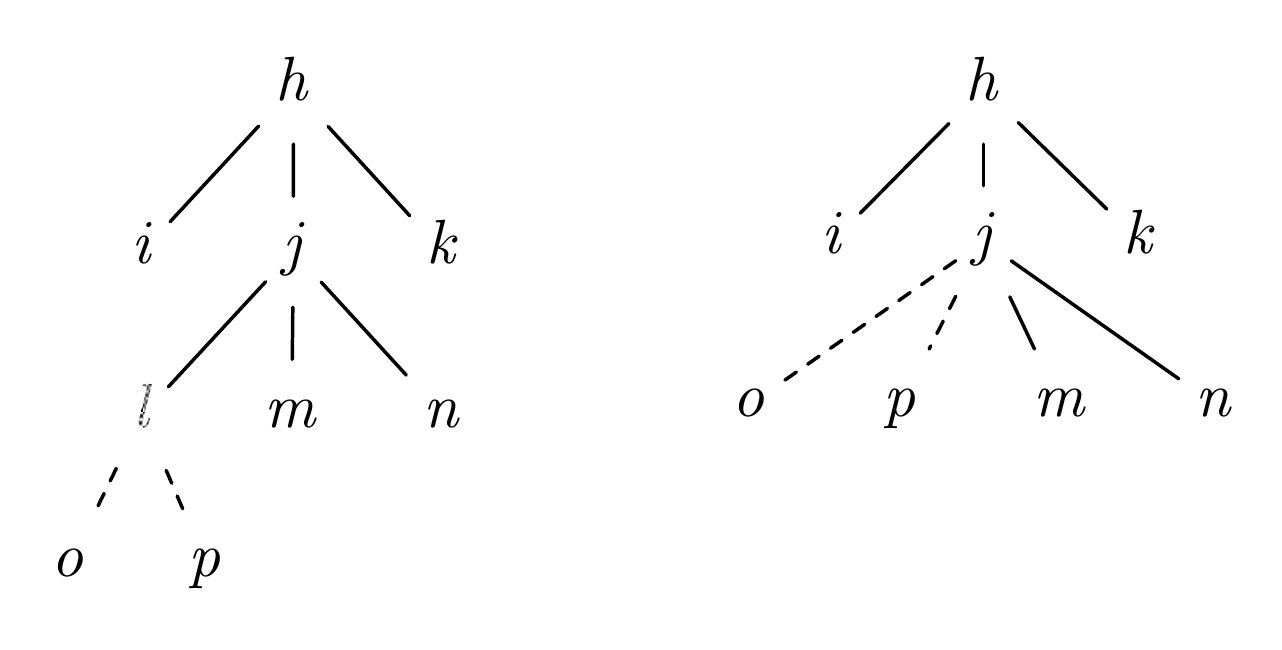}

\label{fig:o1}}

\subfigure[Before and after the nodes $l$ and $j$ have been merged (O2).]{
\raisebox{16mm}{$Merge(j, l, T): \left \{ \begin{array}{l} Children({j+l}) = \{o, p, m, n\}\\ Children(h) = \{i,{j+l},k\} \end{array} \right.$}
\includegraphics[width=\examplewidth]{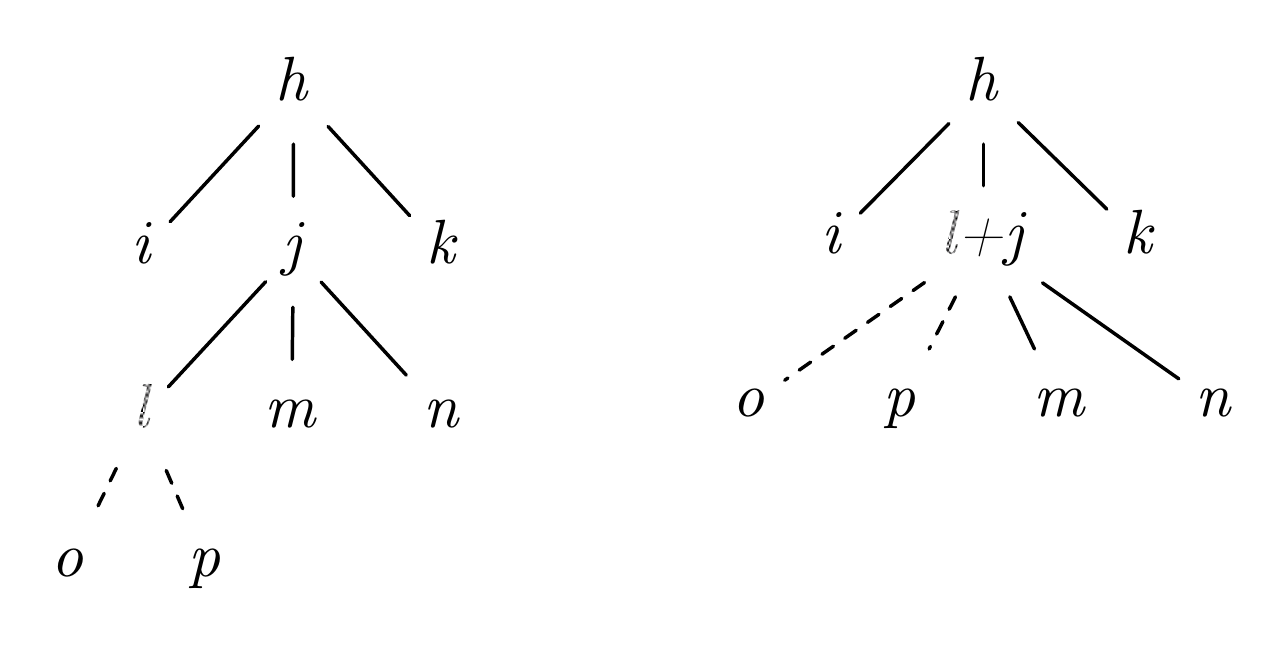} \label{fig:o2}}
\subfigure[Before and after the nodes $l$ and $j$ have been swapped (O3). ]{
\raisebox{20mm}{$Swap(j, l, T): \left \{ \begin{array}{l} Children(j) = \{m, n\}\\ Children(h) = \{i,l,k\} \\ Children(l) = \{o, p, j\}\end{array} \right.$} 
\includegraphics[width=\examplewidth]{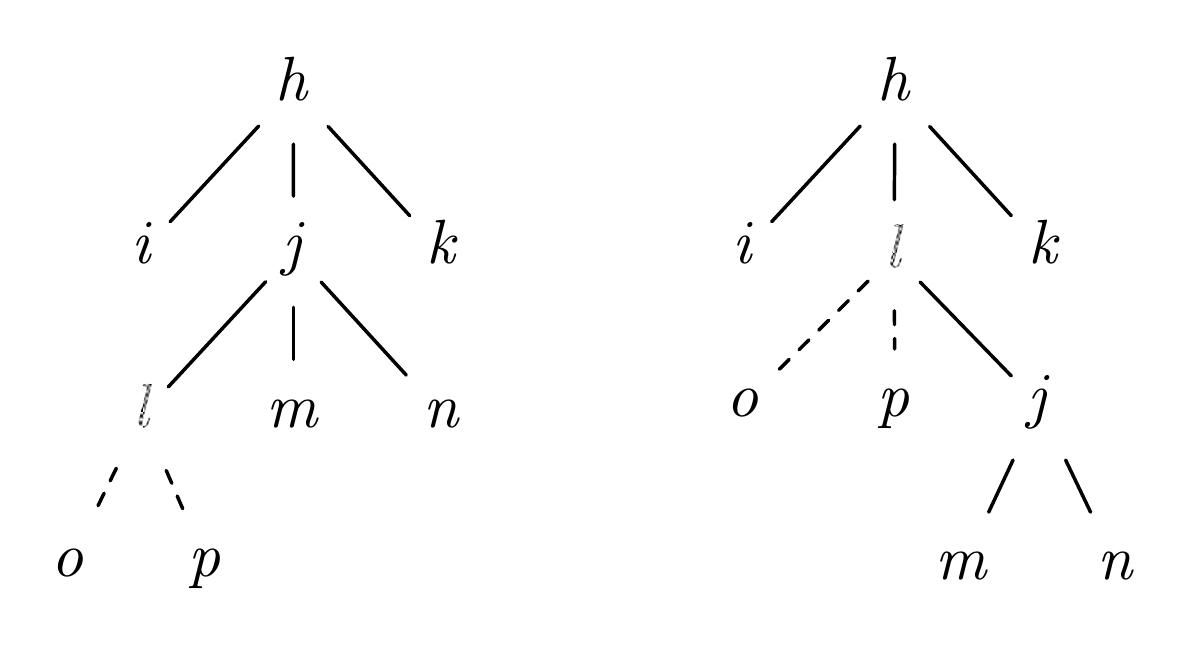} \label{fig:o3}}
\caption{Trees showing the results of the operations defined in O1--O3. $Children(w)$ returns the set of words that depend on $w$. Here we show the value of $Children(node)$ after the operations only if its value is changed by the operations.}
\end{figure}

\renewcommand{\subfigwidth}{0.6\columnwidth}

\renewcommand{\labelenumi}{O\arabic{enumi}.}

\newcommand{\Algo}[2]{\hrule\vspace{0.25em}#1\vspace{0.25em}\hrule\vspace{-0.8em}#2\vspace{-0.8em}}

\subsection{Calculating Tree Matches After Applying Operations}
\label{sec:apply-operations}

The operations O1--O3 are proposed to handle common divergence cases in C1--C3. 
To measure how common C1--C3 is in a language pair, we 
design an algorithm that transforms a tree pair based on 
a word alignment.

The algorithm takes a tree pair $(F, E)$ and a word alignment $A$
as input and creates a modified tree pair 
($S^\prime$, $T^\prime$) and an updated word alignment $A^\prime$ as output. 
It has several steps. 
First, spontaneous nodes (nodes that do not align to any node on the other tree)
are removed from each tree. 
Next, if a node and its parent align to the same node on 
the other tree, they are merged and the word alignment is 
changed accordingly.
%
Finally, the swap operation is applied to a node $f_i$ and its
parent $f_p$ in one tree if they align to $e_i$ and $e_p$ respectively and 
$e_p$ is a child of $e_i$ in the other tree. 
The pseudocode of the algorithm is shown in
Algorithm \ref{tab:alg1}.

\newcommand{\settab}{\hspace{1.5em}\=}

\begin{algorithm}
\small
\SetKwInOut{Input}{input}
\SetKwInOut{Output}{output}

\Input{$c = \big(F,E,A\big)$ \tcp*{A parallel sentence with alignment} }
\Output{$c^{\prime} = \big(F^{\prime}, E^{\prime}, A^{\prime}\big)$ \tcp*{modified output sentence.}}
{\bf Let} $F = (W_F, T_F)$ \;
{\bf Let} $E = (W_E, T_E)$ \;
{\bf Let} $A = \{(f_i, e_j), \ldots, (f_k, e_l)\}$ \;
\Begin{
	\tcp{Step 1(a): Remove spontaneous nodes from $F$}
	\ForEach{$f_i \in W_F$}{
		\If{$\nexists\ e_j : \big( f_i, e_j\big) \in A$}{
			$T_F = Remove(f_i, T_F)$ \tcp*{See Algorithm \ref{alg:remove}}
		}
	}
	\tcp{Step 1(b): Remove spontaneous nodes from $E$}
	\ForEach{$e_j \in W_E$}{
		\If{$\nexists\ f_i : \big( f_i, e_j\big) \in A$}{
			$T_E = Remove(e_i, T_E)$ \tcp*{See Algorithm \ref{alg:remove}}
		}
	}
	\tcp{Step 2(a): Find nodes to merge in $F$ and merge them}
	\ForEach{$(f_i, e_j) \in A$}{
		{\bf Let} $f_p = parent(f_i, T_F)$ \;
		\If{$(f_p, e_j) \in A$}{
			$T_F = Merge(f_i, f_p, T_F)$ \tcp*{See Algorithm \ref{alg:merge}}
			$A = A - \{(f_i, e_j)\}$ \;
		}
	}
	\tcp{Step 2(b): Find nodes to merge in $E$ and merge them}
	\ForEach{$(f_i, e_j) \in A$}{
		{\bf Let} $e_p = parent(e_j, T_E)$ \;
		\If{$(f_i, e_p) \in A$}{
			$T_E = Merge(e_j, e_p, T_E)$ \tcp*{See Algorithm \ref{alg:merge}}
			$A = A - \{(f_i, e_j)\}$ \;
		}
	}
	\tcp{Step 3: Find nodes to swap in $F$ and swap them}
	\ForEach{$(f_i, e_j) \in A$}{
		{\bf Let} $f_p = parent(f_i, T_F)$ \;
		\If{$\exists\ e_c : e_j = parent(e_c, T_E)$ {\bf and} $(f_p, e_c) \in A$}{
			$T_F = Swap(f_i, f_p, T_F)$ \tcp*{See Algorithm \ref{alg:swap}}
		}
	}
		\Return $(F^{\prime}, E^{\prime}, A^{\prime})$ \;
}
\caption[Alignment-based Tree Altering Algorithm]{Algorithm for altering an aligned tree pair.}
\label{tab:alg1}
\end{algorithm}

Now given a corpus $C$ and word alignment between each sentence pair, we can measure the impact of C1--C3 by comparing $CorpusMatch_{Src\rightarrow Tgt}(C)$ scores before and after applying operations O1--O3. This process can also reveal some patterns of divergence (e.g., what types of nodes are often merged), and the patterns can later be used to enhance existing projection algorithms.

\subsection{Relationship to \cite{Dorr:1994:MTD:203987.203993}}	
\label{sec:dorrtypes}

\cite{Dorr:1994:MTD:203987.203993} lists seven types of divergence for language pairs. While our analysis method is more coarse-grained
than the Lexical Conceptual Structure (LCS) that \citeauthor{Dorr:1994:MTD:203987.203993} proposes, 
it is nonetheless able to capture some of the same cases.

\begin{figure}[bthp]
\center
\parbox[b][29mm][t]{0.3\textwidth}{
\texttt{\begin{tabbing}
Juan suele \hspace{2em} \= ir a casa\\
John tends-to \> go home \\
\textit{``John usually goes home''}
\end{tabbing}}}
\hspace{5mm}
\includegraphics[width=0.5\columnwidth]{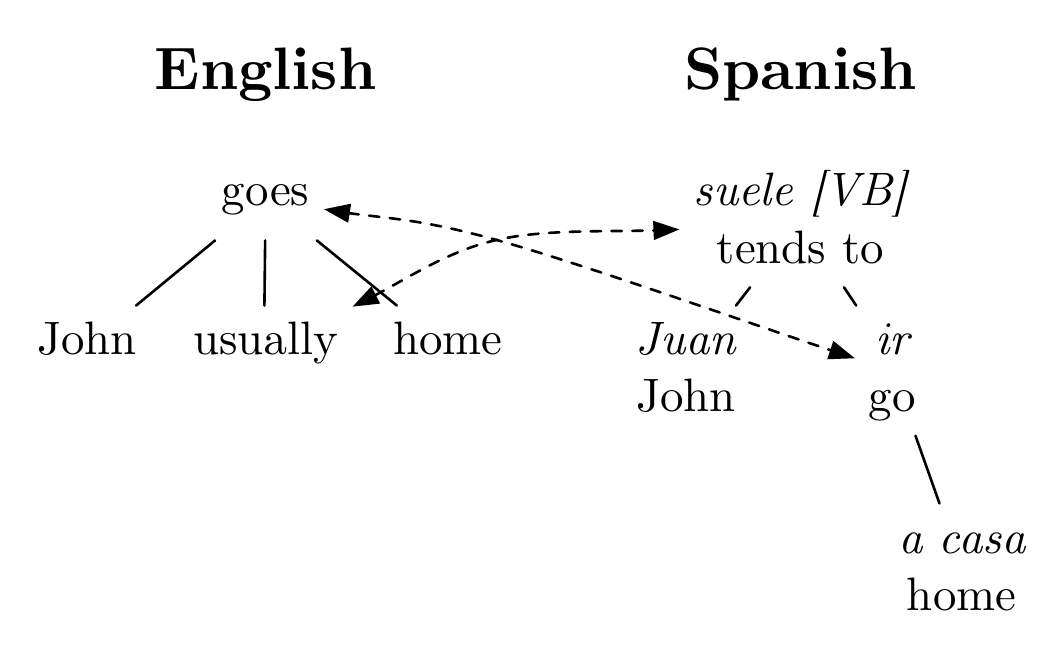}

\caption{An example of \textit{promotional} divergence from \cite{Dorr:1994:MTD:203987.203993}. The reverse in parent-child relation is handled by  the $Swap$ operation. }
\label{fig:promotional}
\end{figure}

For instance, Figure \ref{fig:promotional} illustrates an example 
of what \citeauthor{Dorr:1994:MTD:203987.203993} identified as ``promotional'' divergence, where {\it usually}, a dependent of the verb {\it goes} in English, is ``promoted'' to become the main verb, {\it suele} in Spanish. In this case, 
the direction of the dependency between {\it usually} and {\it goes}
is reversed in Spanish, and thus the {\it swap} operation 
can be applied to the English tree and result in 
a tree that looks very much like the Spanish tree.
 A similar operation is performed for \textit{demotional} divergence cases, such as aligning ``I like eating'' with the German translation \textit{``Ich esse gern''} (``I eat likingly''). Here, the main verb in English (``like'') is \textit{demoted} to an adverbial modifier in German (``\textit{gern}''). The 
{\it swap} operation is applicable to both types of divergence
and treats them equivalently, and so it essentially can
handle a superset of promotional and demotional divergence, 
namely,``head-swapping.''

\begin{figure}[bthp]
\center
\parbox[b][32.5mm][t]{0.3\textwidth}{
\texttt{\begin{Tabbing}
Juan entr\'{o} \hspace{0.8em} \TAB=en la \hspace{.2em} \TAB= casa\\
John entered \TAB> in the \TAB> house \\
\textit{``John entered the house''}
\end{Tabbing}}}\hspace{5mm}
\includegraphics[width=0.5\columnwidth]{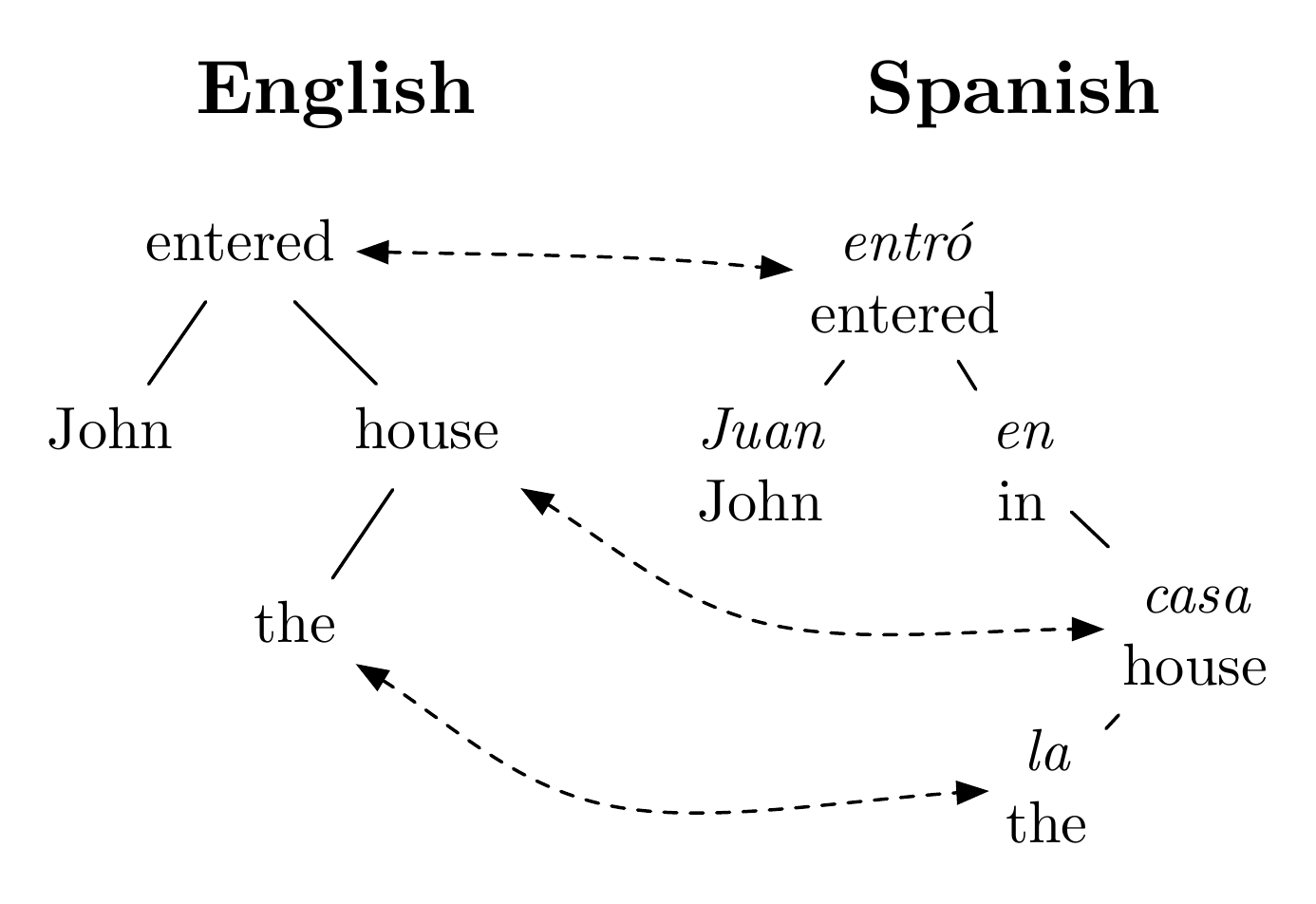}

\caption{Example of structural divergence, which is handled by the \textit{remove} operation.}
\label{fig:structural}
\end{figure}

Another type of divergence that can be captured by our approach is \citeauthor{Dorr:1994:MTD:203987.203993}'s ``structural'' divergence type, as illustrated in Figure \ref{fig:structural}. The difference between the English and Spanish structures in this case is the form of the argument that the verb takes. In English, it is a noun phrase; in Spanish, it is a prepositional phrase. While the tree operations
defined previously do not explicitly recognize
this difference in syntactic labels, the divergence
can be handled by the {\it remove} operation, where
the spontaneous {\it ``en''} in the Spanish side is removed. 

\begin{figure*}[tbhp]
\center

\subfigure[Interlinear text of a sentence pair.]{\begin{minipage}{\textwidth}

\raggedright {\tt 
\begin{tabbing}
mohana \=ne \hspace{1em} \=kala \hspace{3em} \=Arif \=se \hspace{2.5em}\=mInA \=ko \hspace{2em}\=kiwAba \hspace{2em}\=xilavAyI \\
Mohan\>[erg]\>yesterday\>Arif\>[inst]\>Mina\>[dat]\>book\>give-caus\\
\hspace{-1.5em} \textit{\texttt{ ``Mohan caused Mina to be given a book through Arif yesterday.''}}
\end{tabbing}}\vspace{0em}          
\end{minipage}}

\renewcommand{\subfigwidth}{\textwidth}

\subfigure[Initial trees showing spontaneous words on both sides.]{\includegraphics[width=\subfigwidth]{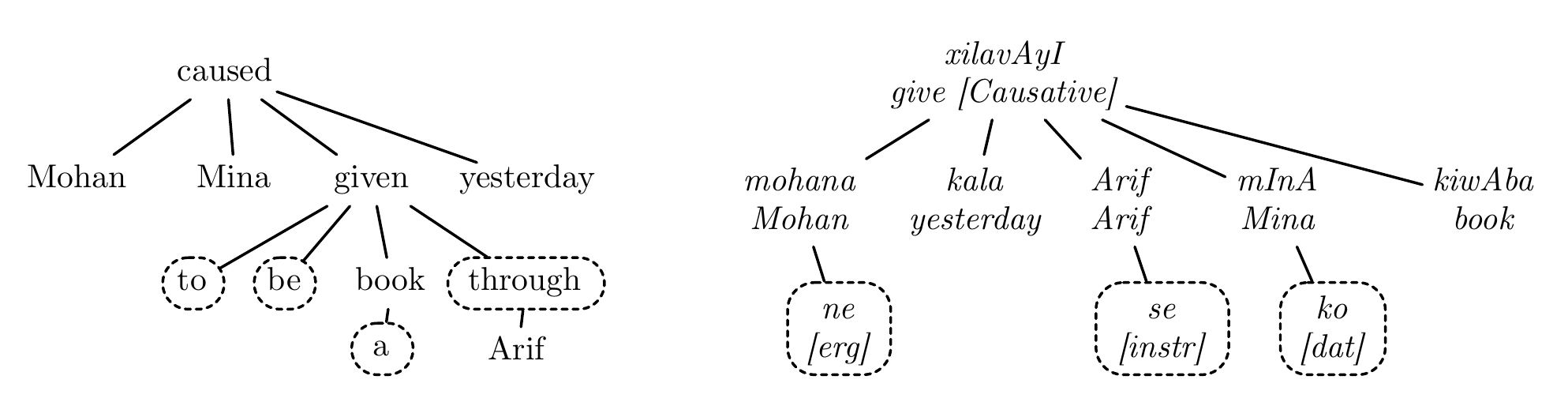}} \\

\subfigure[Altered trees after removing spontaneous words from both sides, and showing conflational divergence between multiple English words and a single Hindi word.]{\includegraphics[width=\subfigwidth]{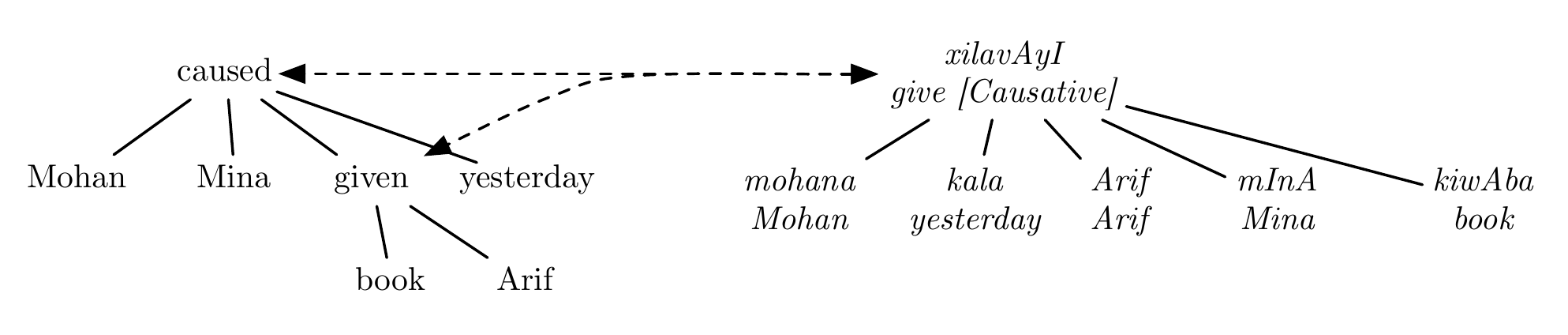} \label{fig:revised-after-spont}} \\

\subfigure[Altered trees after merging multiple words on the English side.]{\includegraphics[width=\subfigwidth]{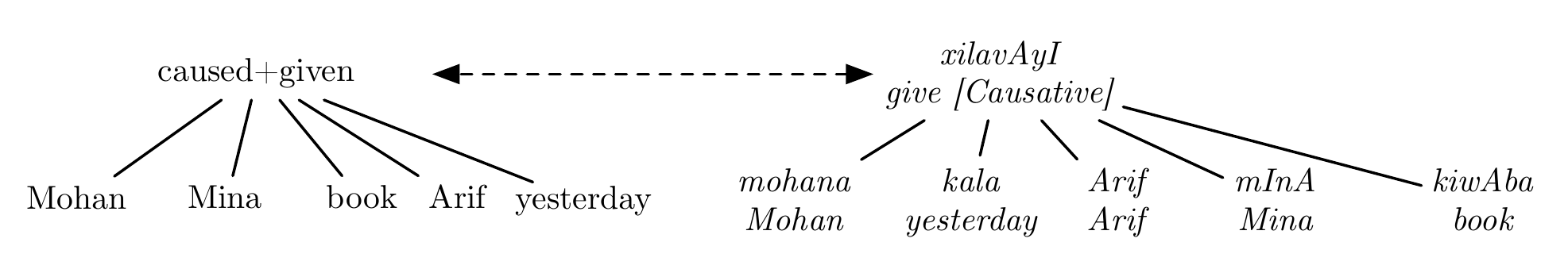} \label{fig:revised-english}}

\caption{Case of conflational divergence, handled by remove and merge operations.}
\label{fig:conflational}
\end{figure*}

Next, \citeauthor{Dorr:1994:MTD:203987.203993}'s description of \textit{conflational} divergence lines up well with the {\it merge} operation (\textit{see Fig \ref{fig:o2}}).  Figure \ref{fig:conflational} illustrates an 
example for English and Hindi, where both sides
have spontaneous words (e.g., {\it to} and {\it a} in English)
and a causative verb in Hindi 
corresponds to multiple verbs in English.  
Figure \ref{fig:conflational}(b) shows the original tree pair,
\ref{fig:conflational}(c) demonstrates the altered tree pair
after removing spontaneous words from both sides. 
Figure \ref{fig:conflational}(d) shows the tree pairs after
the English verbs are merged into a single node.
It is clear that the {\it remove} and {\it merge} operations
make the Hindi and English trees much similar to each other.

 In addition to the four divergence types mentioned above, additional operations could be added to handle other divergence types. For instance, if dependency types (e.g., patient, agent)
 are given in the dependency structure, we can define a new operation that changes the dependency type of an edge to account for \textit{thematic} divergence, where thematic roles are switched as in ``I like Mary'' in English vs. \textit{``Mar\'ia me gusta a m\'i''} (Mary pleases me) in Spanish. Similarly, 
an operation that changes the POS tag of a word can be added to cover \textit{categorial} divergence where words representing the same semantic content have different word categories in the two languages, such as in ``I am hungry'' in English versus \textit{``Ich habe Hunger''} (I have hunger) in German. 


Compared to Dorr's divergence types, whose identification requires
knowledge about the language pairs, our operations on the dependency structure
relies on word alignment and tree pairs and
can be applied automatically.


\subsection{Extending Projection Algorithms}

\label{sec:improve-projection}

Using the techniques described above, and following \cite{Hwa:2004ua}, we can find post-processing rules automatically. Our method couples the divergent cases C1--C3 with corresponding operations O1--O3. As the operations are applied, statistics are kept on the nodes that are affected, and thus common divergence patterns can be detected by analyzing this data. By generalizing  the data found in this analysis, rules that can handle common divergence types could be applied to particular language pairs that exhibit such patterns in the small training corpus.  We will discuss one such attempt at learning a rewrite rule in \S\ref{sec:post-rules}.

\section{Experiments}

With the matching function and tree operations defined
in the previous section,  we looked at a total of eleven language pairs, using the corpora described in Table \ref{tab:data}.

\subsection{Data}
\label{sec:data}

\begin{table}
\centering
\begin{tabular}{|c|c|c|c|}
\hline
Corpus & Language & \# Instances & \# Words (F / E) \\
\hline
Hindi Treebank & Hindi & 147 & 963 / 945\\
\hline
\multirow{7}{*}{ODIN} 
& German & 105 & 747 / 774\\
\cline{2-4}
& Irish & 46 & 252 / 278\\
\cline{2-4}
& Hausa & 77 & 424 / 520\\
\cline{2-4}
& Korean & 103 & 518 / 731\\
\cline{2-4}
& Malagasy & 87 & 489 / 646\\
\cline{2-4}
& Welsh & 53 & 312 / 329\\
\cline{2-4}
& Yaqui & 68 & 350 / 544\\
\hline
\multirow{2}{*}{SMULTRON}
& German & 281 & 6829 / 7236 \\
\cline{2-4}
& Swedish & 281 & 8402 / 9377 \\
\hline

\end{tabular}
\caption[Summary of corpora]{Data set sizes for all languages. For the number of words, the number of words in the foreign (F) language are given first, followed by the number of English (E) words.}
\label{tab:data}
\end{table}

Our work utilizes three corpora for a total of eleven language pairs. The corpora used are the SMULTRON treebank \citep{Smultron2010},
the guideline sentences in IGT form from the Hindi treebank \citep{Bhatt:2009ta}, and several sets of IGT data as used in \citep{Xia:2007wv}. The statistics of the corpora are shown in Table \ref{tab:data}. Ten of the language pairs use English as one side of the language, while the eleventh uses the pair of German and Swedish from the SMULTRON corpus.

In the SMULTRON Treebank, the German and Swedish phrase trees are marked for head children, allowing for the automatic extraction of dependency trees. The English side of the phrase structures do not contain edge labels and are instead converted to dependency trees using a head percolation table \citep{Collins:1999uv}.

From the Hindi Treebank guidelines, we extracted
      example sentences in the form of IGT (i.e., Hindi sentences,
      English gloss, and English translation) and
      the Hindi dependency structures manually created by
      the guideline designers. We obtained dependency structures for the English
      translation by running  the Stanford dependency parser \cite{DeMarneffe:2006uh} and then we hand corrected
      the structures. Word alignment is initially derived from the IGT instances using heuristic alignment following \cite{Xia:2007wv}, and later hand-corrected. The IGT data from \cite{Xia:2007wv} was obtained in the manually corrected dependency forms described in \S\ref{sec:igt}.

\subsection{Match Results}
\label{sec:match-results}

By running Algorithm \ref{alg:matching}, we can calculate the $CorpusMatch_{Src\rightarrow Tgt}$ and $CorpusMatch_{Tgt\rightarrow Src}$ before and after each operation and see how the operation affects the percentage of matched edges in the corpus. As the operations are applied, the percentage of matches between the trees should increase until all the divergence cases that can be handled by operations O1--O3 have been resolved. At this point, the final match percentage can be seen as an estimate of the upper-bound on performance of a simple projection algorithm, if 
C1-C3 can be identified and handled by O1-O3. Table \ref{tab:compare-English-Hindi} shows the full results of this process for English and Hindi, while Table \ref{tab:result-summary} shows a summary for the results in the remaining ten languages.


\begin{table}
\centering
{ \bf English $\leftrightarrow$ Hindi} \vspace{2mm}

\begin{tabular}{|l|cccc|c|}
\hline
& \multicolumn{5}{|c|}{English$\rightarrow$Hindi}  \\
\hline
 &  Match & Swap & Unaligned  & Merge & Edges  \\
\hline 
Baseline & 47.7 & 9.1 & 20.9 & 1.6 & 794  \\ \hline
Remove & 66.1 & 11.7 & 0.0  & 2.1 & 622  \\ \hline
Merge & 69.5 & 12.3 & 0.0  & 0.0 & 586  \\ \hline
Swap & 90.3 & 0.3 & 0.0  & 0.0 & 586  \\ \hline

& \multicolumn{5}{|c|}{Hindi$\rightarrow$English} \\ \hline
Baseline & 46.3 & 5.6 & 20.7  & 1.7 & 816 \\ \hline
Remove & 63.4 & 5.3 & 0.0  & 2.2 & 647 \\ \hline
Merge & 69.2 & 4.6 & 0.0  & 0.0 & 587 \\ \hline
Swap & 89.9 & 2.4 & 0.0  & 0.0 & 587 \\ \hline
\end{tabular}
\caption{Breakdown of edges as operations are applied to the English $\leftrightarrow$ Hindi language pair. The ``Edges'' column represents the number of total edges in the trees of the left hand of the language pair. The numbers given in the other columns are the percentages of those edges that are either in a {\it match}, {\it swap}, or {\it merge} alignment, or the edges for which the child is {\it unaligned}.}
\label{tab:compare-English-Hindi}
\end{table}

\begin{table}
\center
\scriptsize
\begin{tabular}{|c||c|c|c|c|c|c|c||c|c|c|}
\hline
& \multicolumn{7}{|c||}{ODIN DATA} & \multicolumn{3}{c|}{SMULTRON DATA} \\
\hline
& DE & GD & HA & MG & KO & CY & YAQ & DE & SV & DE-SV \\
\hline
Baseline & 76.7 & 72.0 & 54.4 & 57.4 & 56.0 & 75.4 & 54.4 & 40.7 & 37.5 & 43.3\\ \hline
Remove & 93.9 & 87.8 & 95.7 & 88.9 & 88.1 & 95.1 & 90.9 & 63.6 & 62.2 & 73.5\\ \hline
Merge & 95.4 & 92.5 & 97.5 & 97.4 & 95.4 & 97.2 & 95.9 & 71.8 & 73.9 & 82.8\\ \hline
Swap & 96.8 & 94.1 & 97.5 & 98.0 & 96.1 & 98.2 & 96.2 & 83.0 & 84.2 & 87.2\\
\hline
\end{tabular}
\caption{Summary of the results for the remaining ten language pairs, German (DE), Scots Gaelic (GD), Hausa (HA), Malagasy (MG), Korean (KO), Welsh (CY), Yaqui (YAQ), Swedish (SV) and DE-SV. Except for DE-SV, English is the first language of the pair.}
\label{tab:result-summary}
\end{table}


\begin{algorithm}
\scriptsize
\SetKwInOut{Input}{input}
\SetKwInOut{Output}{output}
\Input{A corpus $C$}
\Output{$CorpusMatch_{Src\rightarrow Tgt}(C)$ \\ \hspace{1mm}$CorpusMatch_{Tgt\rightarrow Src}(C)$}
\Begin{
	{\bf Let} $F\rightarrow E$ matches $= 0$ \;
	{\bf Let} $E\rightarrow F$ matches $= 0$ \;
\ForEach{$(F,E,A) \in C$}{
{\bf Let} $F = (W_F, T_F)$ \;
{\bf Let} $E = (W_E, T_E)$ \;
{\bf Let} $A = \{(f_i, e_j), \ldots, (f_k, e_l)\}$ \;
%
%
	\tcp{Get matches for $F\rightarrow E$}
	\ForEach{$\big(f_c, f_p\big) \in T_F$}{
		\tcc{If the child and parent in this edge align with the child$\rightarrow$parent edge of the other tree...}
		\If{$\exists\ e_c, e_p : e_p = parent(e_c, T_E)$ \\
		\hspace{4mm}{\bf and } $(f_p, e_p) \in A$ \\
		\hspace{4mm}{\bf and } $(f_c, e_c) \in A$ \\ }{
			\tcp{Increase the match count.}
			$\big( F\rightarrow E$ matches$\big)$++\;
		}
	}
	\tcp{Get matches for $E\rightarrow F$}
	\ForEach{$\big(e_c, e_p\big) \in T_E$}{
		\tcc{If the child and parent in this edge align with the child$\rightarrow$parent edge of the other tree...}
		\If{$\exists\ f_c, f_p : f_p = parent(f_c, T_F)$ \\
		\hspace{4mm}{\bf and } $(f_p, e_p) \in A$ \\
		\hspace{4mm}{\bf and } $(f_c, e_c) \in A$ \\ }{
			\tcp{Increase the match count.}
			$\big( E\rightarrow F$ matches$\big)$++\;
		}
	}
}
	\Return{$Match(F \rightarrow E) = 100 \times \frac{\big( F\rightarrow E \textrm{matches}\big)}{|T_F|}$} \;
	\Return{$Match(E \rightarrow F) = 100 \times \frac{\big( E\rightarrow F \textrm{matches}\big)}{|T_E|}$} \;
}	
\caption[Match Scoring Algorithm]{Calculating the percentage of matched edges in a corpus $C$.}
\label{alg:matching}
\end{algorithm}

\subsection{Operation Breakdown By POS}
After performing the operations as seen in \S\ref{sec:match-results}, we can get further insight into what precisely is happening within each language by breaking down the operations by the POS tags on which the operations apply. Table \ref{tab:merge-stats} shows some of these POS tag breakdowns for a number of languages, and the frequency with which the given operation applies to the POS tag or POS tag pair out of all the times it is seen in that language. Table \ref{tab:merge-stats} shows phenomena from the language pairs that we would hope to see. For instance, rows 6 and 7 show the English$\rightarrow$German pair merging many nouns as multiple English words are expressed as compounds in German. In another case, Row 8 shows that all Hindi Nouns undergo {\it swap} with prepositions, as Hindi uses postpositions. Noticing this extremely high regularity leads us to the next experiment, where we examine how such regularly-occurring rules might be harnessed to improve projection.
\begin{table}

\center

\begin{tabular}{|c|c|c|c|c|}

\hline
\multicolumn{5}{|c|}{\textbf{Merges}} \\
\hline

Lang Pair& Row \# & \begin{tabular}{c}Child\\POS\end{tabular} & \begin{tabular}{c}Parent\\POS\end{tabular} & \% Merge \\
\hline
\multirow{2}{*}{Eng$\rightarrow$Hin} & 1 & MD & VB & 42.9\% \\
\cline{2-5}
& 2 &NN & NN & 14.3\% \\
\hline

\multirow{2}{*}{Hin$\rightarrow$Eng} & 3 &VAUX &  VM & 45.4\% \\
\cline{2-5}
& 4 &NN & VM & 5.5\% \\
\hline

\multirow{3}{*}{Eng$\rightarrow$Ger} & 5 &NN & NNS & 66.7\% \\
\cline{2-5}
& 6 &NN & NN & 65.4\% \\ \cline{2-5}
& 7 &NNS & NN & 0\% \\

\hline
\multicolumn{5}{|c|}{\textbf{Swaps}} \\
\hline
\multirow{2}{*}{Hin$\rightarrow$Eng} & 8 & NN  & IN & 100\% \\
\cline{2-5}
& 9 & NNP & IN & 20.0\% \\
\hline

\multirow{2}{*}{Ger$\rightarrow$Eng} & 10 & NN & APPRART & 72.7\% \\
\cline{2-5}
& 11 & CC & NN & 61.5\% \\
\hline

\multicolumn{5}{|c|}{\textbf{Removals}} \\
\hline
\multirow{2}{*}{Eng$\rightarrow$Hin} & 12 & DT &  & 86.4\% \\
\cline{2-5}
& 13 & TO &  & 75.6\% \\
\hline

\multirow{2}{*}{Hin$\rightarrow$Eng} & 14 & PSP &  & 69.8\% \\
\cline{2-5}
& 15 & VAUX &  & 18.6\% \\
\hline

\multirow{2}{*}{Eng$\rightarrow$Ger} & 16 & POS &  & 57.1\% \\
\cline{2-5}
& 17 & DT &  & 20.2\% \\
\hline

\multirow{2}{*}{Ger$\rightarrow$Eng} & 18 & PRF &  & 85.2\% \\
\cline{2-5}
& 19 & ADV &  & 43.9\% \\
\hline

\end{tabular}
\caption{Breakdown of significant merge and swap statistics for various language pairs, where the language to the left of the arrow is the one being altered.}
\label{tab:merge-stats}
\end{table}

\subsection{Analyzing Trees For Post-Processing Rules}
\label{sec:post-rules}

Given that we can break down the operations as they apply to particular projected POS tag sequences, we also would like to see if the detected patterns can be used after projection, to correct the projected trees. In Table \ref{tab:swap-percent}, 80\% of the Hindi data was used to train a small corpus on swap patterns were detected. The projection algorithm was run on the remaining 20\%. The first row of Table \ref{tab:threshold} shows the results of the baseline projection algorithm.

 Next, we used the dependencies as projected into Hindi from the English trees as shown in Table \ref{tab:swap-percent} and automatically selected those that occurred with 80\% or higher frequency. The projection algorithm was run as usual, and then every dependency that matched the given pattern was swapped to correct the trees. The second row of the table shows that applying all rules above this threshold actually brought the $F_1$ score down from 64.95 to 63.59. To ensure that only swaps which were both regular and frequent occurred, we tightened the restriction so that only rules that occurred three or more times were chosen. This resulted in an improvement to an $F_1$ score of 73.35 as shown in the final row of Table \ref{tab:threshold}.

\begin{table}
\center
\begin{tabular}{|c|c|c|}
\hline
Projected Dependency & Swap \% & \# Dependencies \\
\hline
NN$\rightarrow$IN & 100\% & 33 \\
\hline
NNP$\rightarrow$IN & 100\% & 12 \\
\hline
JJ$\rightarrow$VBD	& 100\% &	1\\
\hline
VBZ$\rightarrow$VBG & 60\% & 5 \\
\hline
VBG$\rightarrow$VBZ & 28.57\% & 7 \\
\hline
VBG$\rightarrow$NN & 5\% & 20 \\
\hline
\end{tabular}
\caption{Sampling of dependency links which were projected from English onto Hindi. The ``Swap \%'' measures the number of times that particular dependency was found in a swapped alignment with the English tree.}
\label{tab:swap-percent}
\end{table}

\begin{table}
\center 
{\bf Projection Accuracy for English$\rightarrow$Hindi}\vspace{1mm}
\begin{tabular}{|c|c||c|c|c|c|}
\hline
Threshold \% & Count & Precision & Recall & F$_1$ \\
\hline
N/A & N/A & 65.64 & 64.28 & 64.95 \\
\hline
80\% & 0 & 64.26 & 62.93 & 63.59 \\
\hline
80\% & 3 & 74.13 & 72.59 & 73.35 \\
\hline

\end{tabular}
\caption{Results of automatically detecting and applying projected swaps between English and Hindi. The threshold \% describes the rate at which a rule must apply to the POS tag pair before it is automatically applied, and the Count column describes how many times the rule must occur before it is considered. The first row is the baseline with no rules used. }
\label{tab:threshold}
\end{table}

\subsection{Remaining Cases}
\label{sec:remaining-cases}

\begin{figure}[bhtp]
\center

\parbox[b][29.5mm][t]{0.3\textwidth}{
\texttt{\begin{tabbing}
merA \=mana amarIkA meM \=hE \\
my \> mind America in \>is \\
\textit{``I am mentally in America''}
\end{tabbing}}}
\hspace{5mm}
\includegraphics[width=0.5\columnwidth]{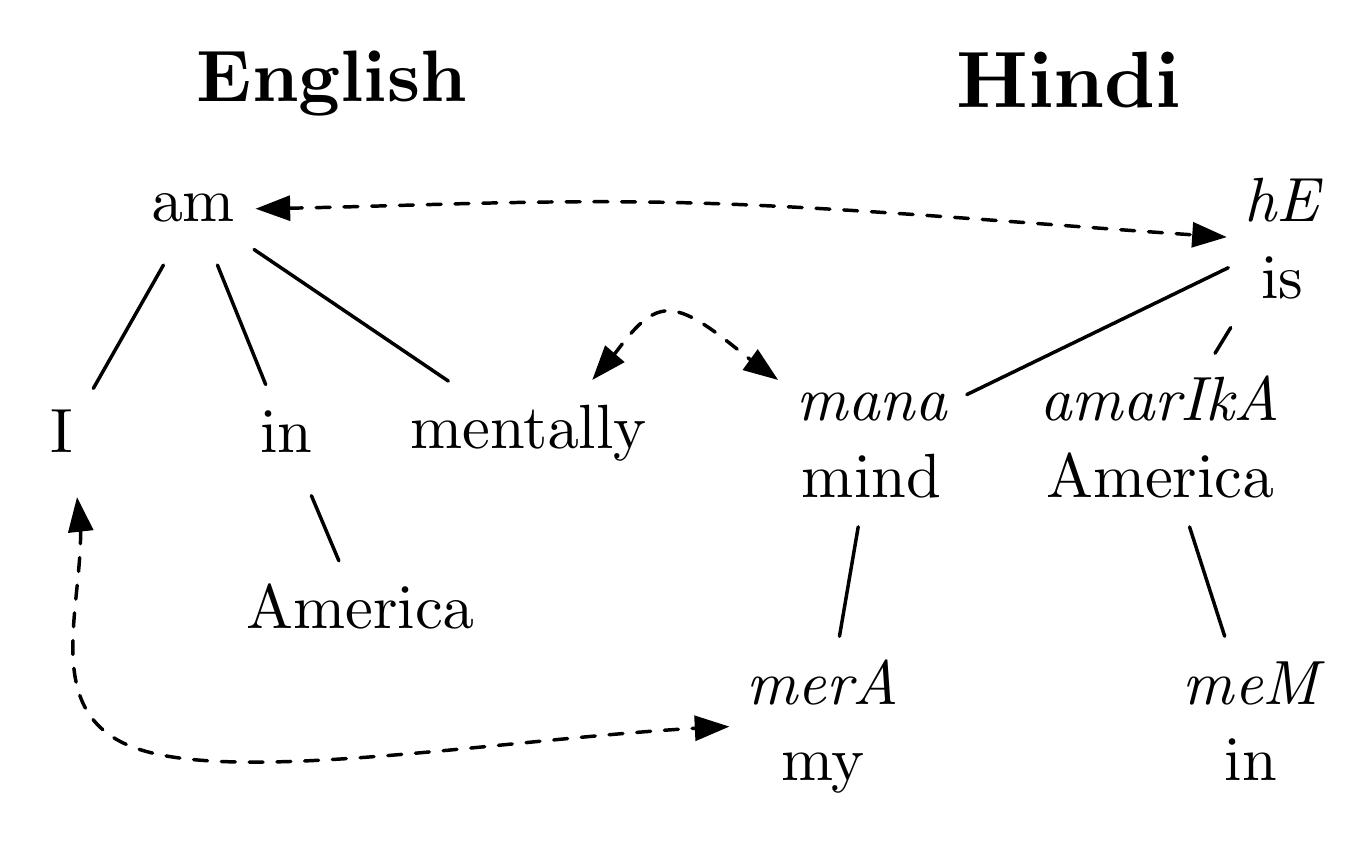}

\caption{A tree pair that still has unmatched edges after applying the algorithm in Table \ref{tab:alg1}. The dotted line indicates word alignment that would be needed to resolve the divergence with the {\it extended} merge operation.}

\label{fig:remaining-case-2}
\end{figure}

After applying three operations, there may still be unmatched edges. An example is given in Figure \ref{fig:remaining-case-2}.
\footnote{It is a topic of debate whether {\it mentally} in English should 
depend on {\it in} or {\it am}. If it depends on {\it in}, handling the 
divergence would be more difficult.}
The dependency edge {\it (in, America)} can be reversed by 
the {\it swap} operation to match the Hindi counterpart.
The difficult part is the adverb {\it mentally} in English 
corresponds to the noun {\it mana (mind)} in Hindi.
If the word alignment includes the three word pairs
as indicated by the dotted lines, one potential way to handle
this kind of divergence is to extend the definition of {\it merge}
to allow edges to be merged on both sides simultaneously -- in this case,
merging {\it am} and {\it mentally} in the English side, and 
{\it hE (is)} and {\it mana (mind)} on the Hindi side.


\section{Discussion of Results}

The results of the experiments above show that the match scoring that we have introduced here has the potential to address many interesting issues arising between language pairs. In this section, we highlight some interesting observations based on the experimental results.

\subsection{Match Scores}

The results of tables \ref{tab:compare-English-Hindi} and \ref{tab:result-summary} are interesting in comparing similarity both across languages and corpora. For instance, in the scores for the baseline ODIN data, we see that the baseline for matches between English and German is the highest out of all the pairs at 76.7\%. Scots Gaelic and Welsh are 72\% and 75.4\%, respectively.  Hausa, Malagasy, Korean, and Yaqui all show baseline scores between 54--57\%. This seems in line with what we would expect, with German and the Celtic languages being closely related to English, and the others being quite unrelated.

Another stark contrast can be seen between all the languages in the ODIN data and the languages in the SMULTRON corpus. While the ODIN sentences tend to be short sentences used primarily for illustrative purposes, the SMULTRON corpus consists of economic, literary, and weather domains. As Table \ref{tab:data} shows, the SMULTRON sentences are much longer on average. A closer look at the SMULTRON translations also shows them to be much freer translations than those found in the ODIN data. While the size of the data sets used here are very small, and the ODIN IGT data may be biased towards illustrative purposes (described as the ``IGT Bias'' in \cite{Xia:2007wv}), it would appear that these results illustrate that the match detection is capable of two interesting corpus analyses. First, by comparing baselines match results among comparable corpora, basic similarities between languages appear to pattern as expected. Second, the freer translations in the SMULTRON data appear with lower scores all across.

One final item of interest from the match results can be seen in the Hindi data in Table \ref{tab:compare-English-Hindi}. Here, there appears to be a large jump 
after the {\it swap} operation has been performed.
Seeing as swapping such as this would be problematic for projection algorithms, this is the inspiration for automatically inferring the swap rules in \S\ref{sec:post-rules}.

\subsection{POS Breakdowns}
\label{sec:pos-breakdown}

\label{sec:patterns}

The breakdown of the operations by language and POS in Table \ref{tab:merge-stats} provides a good opportunity to check that the defined operations conform with  expectations for specific languages. 

For instance, Row 1 in Table \ref{tab:merge-stats} shows Modals (MD) merging with a parent (VB). This is in line with instances such as Figure \ref{fig:revised-after-spont} where Hindi  combines aspect with a verb that is typically expressed as a separate word in English. This does not appear to be a very frequent occurrence, however, as it only occurs for 42.9\% of MD$\rightarrow$VB dependencies.

Row 3, going from Hindi to English shows the case where auxiliary verbs VAUX merge with main verbs VM. These cases typically represent those where Hindi represents tense as an auxiliary verb, whereas English tense is expressed by inflection on the verb.

With regard to spontaneous words in English and Hindi, Row 14 shows that 69.8\% of case markers (PSP) were removed from Hindi that were either absent in English or applied as inflections to the noun,
while 86\% of determiners in English were removed, 
as they are not seen in Hindi (Row 12).

Examining the English and German data in Table \ref{tab:merge-stats}, we first see in Row 5 that 66.7\% of NN-NNS dependencies in English merge. This, along with the 65.4\% of NN-NN dependencies merging, is something we would expect to see in German, as it compounds nouns with far more frequency than English. Interestingly, as row 7 shows, a plural noun child never merges with a parent noun.

Finally, looking more closely at the swaps, we see a surprising 100\% of NN$\rightarrow$IN dependencies are swapped in Hindi, giving further impetus for the rules as described in \S\ref{sec:post-rules}.

\subsection{Automatically Created Rules}

As Table \ref{tab:threshold} illustrates, the potential for automatically created post-processing rules for a projection algorithm is clearly shown by the 13\% increase in $F_1$ score from the uncorrected projection algorithm. While this example focuses only on the easily correctable swap operation, the success here suggests that a similar analysis of the merge operations could be used to correct the 1-to-many projections from English onto other languages that \cite{Yarowsky:2001dl} cited as problematic.

\subsection{Discussion of Issues}

Two large issues that our methodology faces are  data sparsity and translation quality of the sentence pairs in the data sets. The former is somewhat inevitable given the task---a reasonable amount of annotated data is not always likely to exist for languages with scarce electronic resources, and guaranteeing coverage is difficult. As with the Hindi data, however, using IGT as a resource has convenience in both covering wide varieties of phenomena in a language, and providing a gloss that assists in creating word-level alignments. Creating dependency annotation on a small set of data from a source like ODIN \citep{Lewis:2006uv} can get a lot of mileage with a small amount of investment.

Perhaps the more challenging issue is determining whether divergence in a language pair is caused by fundamental differences between the languages, or simply stylistic choices in translation. The latter of these scenarios appeared to be common in portions of the SMULTRON data, where translations appeared to be geared toward naturalness in the target language; in contrast,
the translations in the Hindi guideline sentences
were intended to be as literal as possible. Again, IGT provides a good possible solution, as such examples are often intended specifically for illustrative purposes.
	
\section{Conclusion and Future Work}

In this paper, we have demonstrated a generalizable approach to detecting patterns of structural divergence across language pairs using simple tree operations based on word alignment. We have shown that this methodology can be used to detect similarities between languages on a course level, as well as serve as a general measure of similarity between dependency corpora. Finally, we showed that harnessing these detection methods has potential for improving projection algorithms with little to no expert involvement.

This work further shows that there is still plenty of room for improvement in existing projection methods. In future work, we plan to investigate further ways of learning post-processing rules in projected trees, such as how to properly correct for 1-to-many alignments, and methods for reattaching spontaneous words in the foreign language. In future work, we would also like to examine how labeled dependency edges may play a role in describing divergences between languages, and how these might be more adequately corrected for.

Finally, while \cite{Georgi:2012ys} demonstrated that projected trees can be improved upon by using the projected edges as features in a statistical parser, we would like to follow up on this work by examining how the alignment types outlined in this paper might be also used as features in a parser to further improve dependency parsing for resource-poor languages.

 The techniques described here are promising for maximizing the effectiveness of existing resources such as IGT for languages with little more available. While the amount of electronic resources continues to increase for those languages in which electronic communication is increasingly common, many of these resource-poor languages are still left behind. Though projection techniques may not ultimately be full replacements for large treebank projects, the ability of these techniques to be rapidly deployed is extremely useful for researchers seeking to experiment with new languages at minimal cost.

\bibliography{lrec}

\end{document}